\documentclass{siamart190516}
\ifpdf
\hypersetup{
  pdftitle={Model-free control of embodied systems},
  pdfauthor={Yusheng Jiao, Feng Ling, Sina Heydari, Nicolas Heess, Josh Merel, Eva Kanso}
}
\fi

\usepackage{amsfonts}
\usepackage{algorithm}
\usepackage{algorithmicx}
\usepackage{algorithm,algcompatible,lipsum}
\usepackage[english]{babel}
\usepackage[utf8]{inputenc}
\usepackage{algpseudocode}
\usepackage{array}
\usepackage{arydshln}
\usepackage{cite}
\usepackage{mathtools}

\begin{document}

\title{
Deep Dive into Model-free Reinforcement Learning for Biological and Robotic Systems: Theory and Practice}

\author{Yusheng Jiao%
\thanks{Department of Aerospace and Mechanical Engineering, University of Southern California, Los Angeles, CA 90089.}
\and Feng Ling$^*$
\and Sina Heydari$^*$
\and Nicolas Heess
\thanks{Google Deep Mind, London}
\and Josh Merel
\thanks{Fauna Robotics, New York City}
\and Eva Kanso$^*$
\thanks{Corresponding author: \email{kanso@usc.edu}, \url{https://sites.usc.edu/kansolab/}}
}

\maketitle

\begin{abstract}
Animals and robots exist in a physical world and must coordinate their bodies to achieve behavioral objectives.  
With recent developments in deep reinforcement learning, it is now possible for scientists and engineers to obtain  sensorimotor strategies (policies) for specific tasks using physically simulated bodies and environments. However, the utility of these methods goes beyond the constraints of a specific task; they offer an exciting framework for understanding the organization of an animal sensorimotor system in connection to its morphology and physical interaction with the environment, as well as for deriving general design rules for sensing and actuation in robotic systems.
Algorithms and code implementing both learning agents and environments are increasingly available, but the basic assumptions and choices that go into the formulation of an embodied feedback control problem using deep reinforcement learning may not be immediately apparent.
Here, we present a concise exposition of the mathematical and algorithmic aspects of model-free reinforcement learning, specifically through the use of \textit{actor-critic} methods, as a tool for investigating the feedback control underlying animal and robotic behavior. 
\end{abstract}


\section{Introduction}
Animals operate in complex worlds that are often \textit{dynamic}.
To survive, an animal must transform a rich stream of sensory information into intelligent actions that serve its goals. Objectives as fundamental as foraging for food, avoiding a predator, or navigating a challenging terrain could involve multiple sensory modalities, control actions, and even hierarchical planning before actions are taken. The study of how these decision processes are enacted at the physiological level, that is, the level of neuronal circuits, remains a challenge for research in neuroscience~\cite{Smith2022,Dickinson1999,Tytell2011,Merel2019a,Madhav2020}. 
Recently, bio-inspired robotic systems and simulated bodies interacting with a physical environment emerged as a tractable approach for investigating the sensing and control mechanisms that produce desired behavior, from locomotion on complex terrains to playing soccer~\cite{Aguilar2016,Li2023,Haarnoja2024}. 
Reinforcement learning (RL) is an optimization framework that is particularly suited for this task, especially in unsteady and dynamically-changing environments~\cite{Gunnarson2021, Singh2023}. Unlike
classic optimization methods~\cite{Todorov2002}, RL seeks optimal control laws that can adapt and change with experience~\cite{Degris2012,Haith2013,Cully2015}.

Here, we aim to explain the theory, assumptions and choices that enter in formulating the problem of learning through interactions with the physical environment. We target researchers in the physical and biological sciences and engineering that are newcomers to reinforcement learning and who want a deeper understanding of the mathematical underpinnings and practical challenges that arise when implementing a learning problem.
There is already excellent introductory material to RL; for example, the classic textbook of Barto and Sutton \cite{Sutton2018} and course notes of David Silver \cite{Silver2015}. Yet, the setting we consider in which we seek to obtain sensorimotor strategies or feedback control laws for physical  bodies, biological or robotic, that behave in a physical environment differs from the traditional focus of RL. Our exposition is broadly consistent with formulations in recent simulation~\cite{Heess2017, Haarnoja2024} and robotic implementation \cite{Duchon1998, Kober2013, Choi2023}.

\begin{figure}
\centering
\includegraphics[scale=1]{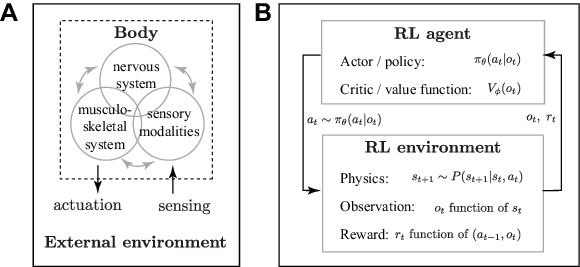}
\caption{\textbf{Reinforcement Learning for Embodied Biological and Robotic Systems:} \textbf{A.} Schematic representation of canonical interactions of a biological organism with its environment. The organism, typically composed of multiple interacting subsystems, potentially with a nervous system and muscles and sensory organs, interacts with the external environment via its physical body: the body acts on the environment and receives sensory feedback including proprioception. \textbf{B.} Reinforcement Learning (RL) provides a framework for studying the interaction of an embodied RL agent with its environment. Modelers decide the type of actions and observations afforded by the embodied agent, as well as the rewards it obtains from the environment. Physical interactions with the environment dictate the time evolution of the state of the system given a choice of action.
The RL agent collects observations $o_t$ and rewards $r_t$ and take actions $a_t$ according to the RL policy $\pi_\theta(a_t | o_t)$.
Model-free RL is a type of learning where no explicit model of the physical transition rules $P(s_{t+1}|s_t,a_t)$ is used or learned by the agent in the process of learning an optimal policy for taking actions $a_t$. In the actor-critic methods, the RL agent is comprised of a policy (actor) $\pi_\theta(a_t|o_t)$ and a value function (critic) $V_\phi(o_t)$, parametrized by $\theta$ and $\phi$, respectively. }
\label{fig:set-up}
\end{figure}

For concreteness, we consider a single \textit{agent}, which we view as an abstract representation of the parts of the animal or robot responsible for sensorimotor decisions. We focus on settings where the agent learns a control law designed to achieve a single predefined \textit{objective} related to a specific task.
We divide the world into a \emph{body} controlled by the agent and through which the agent senses, and an \emph{external environment} that encompasses everything outside of what the agent can control. The actions that control the body are decided in response to sensory observations of both the body itself, \textit{i.e.}, proprioception, and observations of the external environment, \textit{i.e.}, exteroception or telereception; see \cref{fig:set-up}A. 
That the agent is restricted to control and sense through a body is a key characteristic of \textit{embodiment} or \textit{embodied systems}.
The agent's behavior is constrained by its morphology and the physics of its interaction with the environment. 
The loop of an agent observing an environment and acting on it is referred to as the \textit{perception-action cycle}. 

There are some philosophical subtleties in this division of the world into agent, body, and environment~\cite{Pfeifer2006, Barrett2011}. 
While in many animals, the nervous system responsible for control decisions naturally corresponds to the agent, and the musculo-skeletal system along with the sensory modalities is considered the body, the distinction between agent and body and between body and environment is, in essence, a simplification or a modeling choice to be assessed on a case-by-case basis.

In this exposition, 
we outline the structure of embodied reinforcement learning and the underlying mathematical foundation in~\cref{sec:rl_math}. We then explain the theoretical formulation of actor-critic methods in~\cref{sec:Training:AC} and discuss the algorithmic aspects of implementing these methods in~\cref{sec:rl_critic} and~\cref{sec:rl_train}. In ~\cref{sec:implement}, we provide general guidelines for model design.

\section{Mathematical underpinnings of model-free RL}
\label{sec:rl_math}
The RL problem consists of learning a rule or \textit{policy} $\pi(a | o)$ that distates how to select an \textit{action} $a$ based on an \textit{observation} $o$ about the \textit{state} $s$ of the body and environment. The state evolves over time $t$ following the laws of physics, such as the conservation of linear and angular momenta. The policy is probabilistic. It is represented as a conditional probability density function $\pi(a | o)$ of taking an action $a$ given an observation $o$. Such representation considers $a$ as a continuous random variable. For discrete actions, the policy is a conditional probability. The RL agent is expected to learn a policy $\pi$ that maximizes an \textit{objective function} $\mathcal{J}(\pi)$ from repeated interactions with the environment. 

Hereafter, we represent the policy using a parametric function approximator parameterized by $\theta$. The problem of optimizing the control policy $\pi$ becomes that of optimizing its parameters $\theta$. To emphasize this, we use the notation $\pi_\theta$. 

This optimization process depends fundamentally on the interaction between the RL agent and the RL environment; see~\cref{fig:set-up}B. The RL environment includes the time evolution of the state of body and environment. Without loss of generality, and for consistency with most RL text and code, we consider discrete time $t = 0, 1, 2, \ldots$, such as, if the system is in state $s_t$ at time $t$, it evolves to state $s_{t+1}$ at time $t+1$. To emphasize their dependence on time, we denote the state $s_t$, action $a_t$, and observation $o_t$ by the subscript $t$. Additionally, at each time step $t$, the RL environment provides a reward $r_t$ to the RL agent as feedback. 

The time evolution of the system can be described by the probability $P(s_{t+1}|s_t,a_t)$ of transitioning to state $s_{t+1}$ from state $s_t$ given action $a_t$. This transition rule with the RL policy $\pi_\theta(a_t|o_t)$ is  
a Markov Decision Process (MDP) \cite{Bellman1957}. At the new state $s_{t+1}$, new observable features $o_{t+1}$ of the environment are generated. The agent uses the policy $\pi_\theta$ to decide what new action $a_{t+1}$ to take based on the new observations $o_{t+1}$. 
When the agent has access to only partial observations $o_t$ of the state $s_{t}$, the system is called a Partially Observable Markov Decision Process (POMDP) \cite{Aastrom1965,Kaelbling1998,Jaakkola1995,Spaan2012}. POMDPs pose additional challenges to obtaining a control policy because the state of the system may not be uniquely determined from partial observations: different states can produce the same observations, making them indistinguishable for the agent. Yet, almost all control problems in real life belong to the POMDP category~\cite{Reddy2018,Verma2018}. A careful choice of observations is thus crucial for success of the RL policy, and memory of past observations might help. We return to this topic in \cref{sec:implement}.

\begin{figure}
\centering
\includegraphics[scale=1]{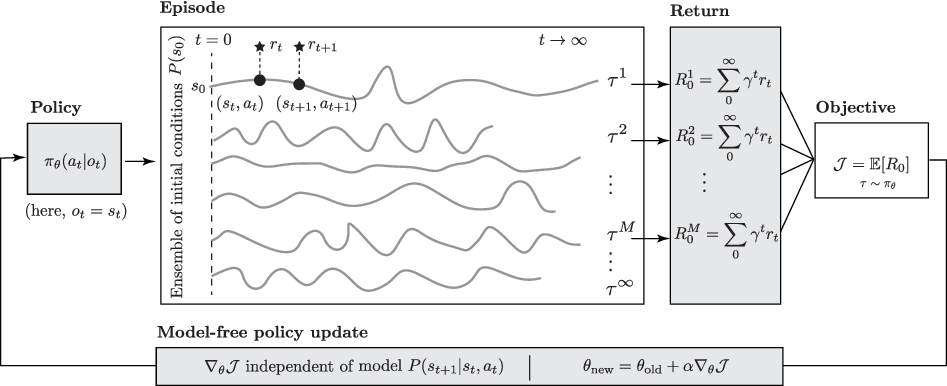}
\caption{\textbf{Reinforcement Learning:}
 RL consists of learning a policy $\pi_\theta(a_t | o_t)$, parameterized by $\theta$, that maximizes an objective function $\mathcal{J}$. Learning is based on repeated interactions with the environment: starting from a rule for choosing an action $a_t$ when in state $s_t$, which could be initially random, the RL agent uses a noisy version of this rule to explore the environment and collects an ensemble of trajectories $\tau\equiv{a_t,o_t}$ and rewards $r_t$, based on which the policy $\pi_\theta(a_t | o_t)$ is updated. Model-free RL describes a framework where the policy update does not depend on an explicit model $P(s_{t+1}|s_t,a_t)$ that represents the interactions of the agent with its environment. For simplicity, we show a policy updated every episode, but in most RL implementations, the update cycle is independent of the choice of episode. 
}
\label{fig:intuition}
\end{figure}

From an initial state $s_0$, the system evolves based on actions $a_t \sim \pi_\theta(a_t|o_t)$ instructed by the policy, while the agent collects observations $o_t$ and rewards $r_t$. This time evolution defines a trajectory $\tau$ of actions, observations, and rewards $\tau = (a_0, o_0, r_0, \ldots, a_t, o_t, r_t, a_{t+1}, o_{t+1}, r_{t+1}, \ldots )$. As $t$ evolves from 0 to $\infty$, we obtain an \textit{infinite time-horizon trajectory}. The infinite time-horizon trajectory is not unique. A different trajectory is realized with each time sequence of the Markovian decision process, even when starting from the same initial condition $s_0$. We thus speak of a distribution $p_\theta(\tau)$ of trajectories $\tau$. We often consider a distribution $P(s_0)$ of initial states and obtain a distribution $p_\theta(\tau)$ as shown in~\cref{fig:intuition}. If the observation $o_t$ is a deterministic function of the state $s_t$, the distribution $p_\theta(\tau)$ of trajectories $\tau$ that are induced jointly by the distribution of initial states $P(s_0)$, the environment dynamics $P(s_{t+1} | s_t, a_t)$, and the policy $\pi_\theta(a_t | o_t)$ can be written as
\begin{equation}
\label{eq:traj_decompose}
     p_\theta(\tau) = P(s_0) \prod_{t=0}^\infty P(s_{t+1}|s_t, a_t) \pi_\theta(a_t|o_t).
\end{equation}

The agent uses the information from trajectories $\tau$ to update the parameters of the policy $\pi_\theta$ in a way that maximizes an objective function or minimizes a cost function $\mathcal{J}(\pi_\theta)$. %
To define $\mathcal{J}$, we introduce the \textit{return} ${R}_t$ in terms of the rewards $r_t$ collected over an infinite-time trajectory $\tau$, 
\begin{equation}
 R_t=\sum_{t'=t}^{\infty} \gamma^{t'-t} r_{t'}.
 \label{eq:return}
 \end{equation}
The parameter $\gamma \in [ 0, 1)$ is a \emph{discount factor}; it determines the preference for future over immediate rewards. The objective function $\mathcal{J}(\pi_\theta)$ is defined by the \emph{expected return} over the distribution 
$p_\theta(\tau)$ of trajectories $\tau$,
\begin{equation}
\label{eq:obj}
    \mathcal{J}(\theta)= %
    \mathbb{E}_{\pi_\theta}\big[R_0\big]
    = \int R_0 p_\theta(\tau)\mathrm{d}\tau.
\end{equation}
Here, the return $R_0 = \sum_{t'=0}^\infty\gamma^{t'} r_{t'}$ is evaluated at $t=0$, and the expectation $\mathbb{E}_{\pi_\theta}\big[R_0\big]$ is taken with respect to a distribution $p_\theta(\tau)$ of trajectories $\tau$. Note that, because $\pi_\theta$ is determined by its parameters $\theta$, we wrote the objective function $\mathcal{J}(\pi_\theta)$ directly in terms of $\theta$.

Evaluating $\mathcal{J}(\theta)$ is analytically and computationally intractable because the return $R_0$ is defined over an infinite time horizon and because the expectation $\mathbb{E}_{\pi_\theta}\big[R_0\big]$ involves evaluating a high-dimensional integral over a distribution $p_\theta(\tau)$ of trajectories. A number of approaches have been proposed to approximately maximize $\mathcal{J}(\theta)$ with respect to $\theta$~\cite{Kimura1995, Baxter2000, Sehgal2019, Rafati2020}. Here, we describe a gradient descent approach that directly approximates the gradient
\begin{equation}
\label{eq:gradJ_temp}
    \nabla_\theta \mathcal{J}(\theta) = \int R_0\nabla_\theta p_\theta(\tau)\mathrm{d}\tau.
\end{equation}
Specifically, we show that $\nabla_\theta  \mathcal{J}$ can be evaluated from sampled trajectories $\tau$ in a \textit{model-free} fashion.
By \textit{model free}, we mean that  the gradient $\nabla_\theta  \mathcal{J}$ can be expressed in a way that is independent of an explicit model of the state transition rule $P(s_{t+1}|s_t,a_t)$.

To approximate $ \nabla_\theta \mathcal{J}(\theta)$ in~\eqref{eq:gradJ_temp}, we need to evaluate $\nabla_\theta p_\theta(\tau)$. We start from the identity $\nabla_\theta  p_\theta =  p_\theta \nabla_\theta \log  p_\theta$, also known as the ``likelihood ratio trick''. According to~\eqref{eq:traj_decompose}, the log-probability can be decomposed to $\log  p_\theta(\tau) = \log P(s_0) + \sum_{t=0}^\infty \left[\log \pi_\theta(a_t | o_t) + \log P(s_{t+1} | s_t,a_t)\right]$, where $P(s_0)$ and $P(s_{t+1} | s_t,a_t)$ do not explicitly depend on $\theta$. We get that
\begin{equation}
    \nabla_\theta p_\theta(\tau) =  p_\theta(\tau)\nabla_\theta\sum_{t=0}^\infty\log \pi_\theta(a_t|o_t).
    \label{eq:pgrad_log}
\end{equation}
Substituting~\eqref{eq:pgrad_log} and~\eqref{eq:return} into~\eqref{eq:gradJ_temp}, we arrive at
\begin{equation}
    \nabla_\theta \mathcal{J}(\theta)
    = \mathbb{E}_{\pi_\theta}\left[\sum_{t=0}^{\infty}\sum_{t'=0}^\infty\gamma^{t'}r_{t'}\nabla_\theta \log\pi_\theta \right].
    \label{eq:gradJ}
\end{equation}
The policy at time $t$ does not have an effect on rewards collected at $t'<t$; indeed, in \cref{sec:appendix:obj}, we provide a 
proof that
$\mathbb{E}_{\pi_\theta}\left[\sum_{t=0}^{\infty}\sum_{t'=0}^{t-1} \gamma^{t'}r_{t'}\nabla_\theta \log\pi_\theta(a_t|o_t) \right] = 0$. 
Substituting this identity into~\eqref{eq:gradJ} and recalling~\eqref{eq:return}, we arrive at the expression
\begin{equation}
    \nabla_\theta \mathcal{J}(\theta)
    = \mathbb{E}_{\pi_\theta}\left[\sum_{t=0}^{\infty}\sum_{t'=t}^\infty\gamma^{t'}r_{t'} \nabla_\theta \log\pi_\theta\right] = \mathbb{E}_{\pi_\theta}\left[\sum_{t=0}^{\infty}\gamma^t R_t\nabla_\theta \log\pi_\theta \right].
    \label{eq:pgrad}
\end{equation}

The gradient in~\cref{eq:pgrad} cannot be readily calculated  in closed form but can be approximated using a Monte Carlo approach by sampling trajectories $\tau^j \sim p_\theta(\tau)$, $j=1,\ldots, M$, and estimating the gradient along these trajectories
\begin{equation}
    \nabla_\theta \mathcal{J}(\theta)
    \approx \frac{1}{M} \sum_{j=1}^M \left[  \sum_{t=0}^{\infty} \gamma^t R_t^j \nabla_\theta \log \pi_\theta(a_t^j|o_t^j) \right].
    \label{eq:mc_pgrad}
\end{equation}
This estimate of $\nabla_\theta \mathcal{J}$ is then used to update the policy parameters
\begin{equation}
\theta_{\textrm{new}}=\theta_{\textrm{old}}+\alpha\nabla_\theta\mathcal{J},\end{equation} 
where $\alpha$ is a constant parameter known as the \textit{learning rate}. 

An important feature of this approach is that, to evaluate $\nabla_\theta \mathcal{J}(\theta)$, explicit knowledge of the environment dynamics is not required. The trajectories $\tau^j$ are sampled from the distribution $p_\theta(\tau)$, which comprises the system dynamics, but $\nabla_\theta \mathcal{J}(\theta)$ does not depend on $P(s_{t+1}|s_t, a_t)$. To evaluate the gradient $\nabla_\theta \mathcal{J}(\theta)$, we merely require the ability to sample trajectories by interacting with the environment --  the gradient update is \textit{model free}.  

This approach for obtaining the gradient without differentiation through a forward model of the environment was popularized and named \text{REINFORCE} in \cite{Williams1992}.
The resulting class of algorithms is named \textit{policy gradient method} \cite{Williams1992, Sutton2000, Kakade2002a, Peters2008, Peters2006} because it computes a gradient of the long-horizon objective with respect to the policy to learn. 
Policy gradient methods are widely used to learn complex control tasks and are often regarded as the most effective reinforcement learning techniques, especially for robotic applications \cite{Peters2006}.
They are scalable to high dimensional continuous state and action problems and are guaranteed to converge to a locally optimal policy \cite{Sutton2000, Pirotta2013}. However, these algorithms are rarely built on the form presented in \cref{eq:mc_pgrad}. The reason is that, without further modification, the gradient estimate has high variance because it is highly sensitive to the randomness along trajectories $\tau^j$ and the choice of the gradient step size. To remedy this, a class of RL methods, known as \textit{actor-critic} methods, replace the sample estimate of the return $R_t^j$ at $s_t$ along a single trajectory $\tau^j$ with the expectation of return for all trajectories starting at $s_t$, as discussed next.

\section{Actor-critic methods}
\label{sec:Training:AC}

Actor-critic methods are designed to improve data efficiency and stability of policy gradient algorithms by re-expressing the estimate $R_t^j$ of the return in~\eqref{eq:mc_pgrad} in terms of an expected return $\mathbb{E}_{\pi_\theta}[R_t|s_t,a_t]$ over all possible future actions.  The expected return $\mathbb{E}_{\pi_\theta}[R_t|s_t]$ over all possible actions is then subtracted from $\mathbb{E}_{\pi_\theta}[R_t|s_t,a_t]$ to get a relative expectation.
This relative expectation is called the \textit{advantage}. 
By definition, the advantage is implemented by introducing two new functions available to the RL agent: $Q(s_t,a_t)$ and $V(s_t)$. Specifically,
\begin{equation}
    \mathcal{A}_t  \coloneqq Q(s_t,a_t) - V(s_t)
    \coloneqq \mathbb{E}_{\pi_\theta} [R_t|s_t,a_t] - \mathbb{E}_{\pi_\theta}[R_t|s_t],
\end{equation}
where
 the \textit{state value function}, or simply \textit{value function}, provides the return, that is, discounted sum of future rewards, that the agent expects to receive from state $s_t$ under policy $\pi_\theta$, 
\begin{equation}
V(s_t) = \mathbb{E}_{\pi_\theta}\big[R_t| s_t] = \mathbb{E}_{\pi_\theta}\left[\sum_{t'=t}^{\infty} \gamma^{t'-t} r_t\bigg| s_t \right],
\label{eq:vval}
\end{equation}
and the \textit{state-action value function}, or simply \textit{$Q$-function}, expresses the expected return assuming that the agent starts from state $s_t$ and selects an action $a_t$ under policy $\pi_\theta$,
\begin{equation}
Q(s_t,a_t) = \mathbb{E}_{\pi_\theta}\big[R_t | s_t, a_t \big] = \mathbb{E}_{\pi_\theta}\left[\sum_{t'=t}^{\infty} \gamma^{t'-t} r_t\bigg | s_t, a_t \right]. 
\label{eq:qval}
\end{equation}
Clearly, $V$ and $Q$ are related via $V(s_t) = \mathbb{E}_{\pi_\theta}[Q(s_t,a_t) | s_t]$. When expressed recursively, these functions are known as the \textit{Bellman optimality equations},  
\begin{equation}
    V(s_t) = \mathbb E_{\pi_\theta}\big[r_t+V(s_{t+1})|s_t\big], \qquad
    Q(s_t,a_t) = \mathbb E_{\pi_\theta}\big[r_t+V(s_{t+1})|s_t,a_t\big].
\end{equation}
The definitions of $V$ and $Q$ %
may be puzzling at first because the RL agent may not have explicit knowledge of the state $s_t$; it only observes $o_t$ of the state $s_t$. For now, to further develop the actor-critic method, we assume that $o_t = s_t$ such that the RL agent has full knowledge of $s_t$. Later, we discuss how to evaluate the $V$ and $Q$ functions when only partial observations about the state are available to the agent.

Given~\cref{eq:vval} and~\cref{eq:qval} and the properties of conditional expectations,  we express the gradient in~\eqref{eq:pgrad} in terms of $V(s_t)$ and $Q(s_t,a_t)$ (derivation provided in \cref{sec:appendix:pg})
\begin{equation}
    \nabla_\theta \mathcal{J}(\theta)
    = \mathbb{E}_{\pi_\theta}\left[ \sum_{t=0}^{\infty} \gamma^t \Bigl(Q(s_t, a_t) - V(s_t)\Bigr) \nabla_\theta \log \pi_\theta(a_t|s_t) \right] = \mathbb{E}_{\pi_\theta}\left[ \sum_{t=0}^{\infty} \gamma^t \mathcal{A}_t \nabla_\theta \log \pi_\theta(a_t|s_t) \right].
    \label{eq:pgrad_values}
\end{equation}
This expression differs from the one in \cref{eq:pgrad} in two important ways. First, rather than ``scoring'' an action directly by the return $R_t$ received along a particular trajectory, we now use the \emph{expected} return, that is, the return averaged over all future trajectories that start at $s_t$ and with action $a_t$. Second, we subtract the expected future return $V(s_t)$ that the policy $\pi_\theta$ would receive when starting at $s_t$. The term $V(s_t)$ acts as a \emph{baseline}: instead of looking at the absolute expected return, we consider how much better a given action is compared to what the policy would receive on ``average''. Thus, the functions $Q(s_t,a_t)$ and $V(s_t)$ play the respective roles of scoring actions and acting as a reference. It is because of this interpretation that the difference $\mathcal{A}(s_t,a_t) = Q(s_t,a_t)-V(s_t)$ is called the \emph{advantage}.

Although in principle \cref{eq:pgrad} and \cref{eq:pgrad_values} express the same quantity in two different ways, 
when used as part of a Monte Carlo estimate, the modification in \cref{eq:pgrad_values} is significant. Indeed, given $Q$ and $V$, the Monte Carlo estimate of~\cref{eq:pgrad_values} is given by ($\mathcal{A}_t^j = Q(s_t^j,a_t^j) - V(s_t^j)$)
\begin{equation}
    \nabla_\theta \mathcal{J}(\theta)
    \approx \frac{1}{M} \sum_{j=1}^M \left[  \sum_{t=0}^{\infty} \gamma^t \mathcal{A}_t^j
     \nabla_\theta \log \pi_\theta(a_t^j|s_t^j) \right].
    \label{eq:mc_pgrad_values}
\end{equation}
This estimate has significantly lower variance than the estimate in \cref{eq:mc_pgrad} for the same two reasons mentioned above: the sample estimate $R^j_t$, which is affected by the randomness along the trajectory $\tau^j$ following $s_t^j$, is replaced with the expectation, and the baseline is subtracted. The absolute value $Q(s_t,a_t)$ for a given action has little meaning; what matters is the value of an action $a_t$ \emph{relative} to other possible choices of action in that same state.

\section{Approximating the critic}
\label{sec:rl_critic}
How are $Q(s_t,a_t)$ and $V(s_t)$ computed? 
This is a difficult problem in itself and is typically intractable analytically. 
These functions are accessible via approximations that replace $V$ and $Q$ by the estimates $\widehat{V}$ and $\widehat{Q}$. A common practice is to use a function approximator $\widehat V (s_t) = V_\phi(s_t)$, parameterized by $\phi$, for the value function and to use a Monte Carlo estimate for the Q-function. For example, considering a single-trajectory Monte Carlo estimate of the $Q$-function, we get $\widehat{Q}(s_t,a_t) = R_t$. In practice, we only have access to finite time-horizon trajectories, say, $t=0,\ldots, T$ and only the rewards $r_t$ from $t$ to $T$ are available towards computing $R_t$. Thus, $R_t$ is typically truncated over $K$ time steps $(K\le T-t+1)$, such that the Q-function estimate follows a $K$-step truncated approximation
\begin{equation}
    \widehat{Q}(s_t,a_t) = r_t + \gamma r_{t+1} + \gamma^2 r_{t+2} +\ldots + \gamma^{K-1}r_{t+K-1}.
    \label{eq:qvalKstep}
\end{equation}
The residual, that is, the truncated portion, is $\sum^\infty_{t'=t+K}\gamma^{t'-t}r_{t'}$. The quality of the $K$-step approximation depends on $K$ and $\gamma$: if $K$ is large and $\gamma$ is small, the influence of the residual may be small. But we can do better. Instead of neglecting the residual contribution to the return beyond $K$ steps, we can substitute the residual with its expectation $\mathbb E_{\pi_\theta}[\sum^\infty_{t'=t+K}\gamma^{t'-t}r_{t'}|s_t]$, which by definition~\cref{eq:vval}, is equal to $\gamma^K V_\phi(s_{t+K})$. This leads to a $K$-step bootstrapped version of $\widehat{Q}$ given by
\begin{equation}
    \widehat{Q}(s_t,a_t) = r_t + \gamma r_{t+1} + \gamma^2 r_{t+2} +\ldots + \gamma^{K-1}r_{t+K-1} + \gamma^K V_\phi(s_{t+K}).
   \label{eq:hatQK} 
\end{equation}

How is $V_\phi(s_{t})$ approximated? Given a specific class of function approximators, say neural networks, the parameters $\phi$ of the function approximator $V_\phi(s_{t})$ are computed by minimizing a loss function $L(\phi)$ equal to the mean-square error between $V_\phi(s_{t})$ and a Monte Carlo estimate $\widehat{V}(s_t)$ based on available trajectories.
\begin{align}
    L(\phi) = E_{\pi_\theta} \left [\sum_t \left(V_\phi(s_t) - \widehat{V}(s_t) \right )^2 \right].
    \label{eq:bellman_loss}
\end{align}
Although $\widehat{V}(s_t)$ may depend on $\phi$, we typically do not consider  its contribution to the gradient 
\begin{equation}
\nabla_\phi {L}(\phi) = E_{\pi_\theta} \left [\sum_t \left(V_\phi(s_t) - \widehat{V}(s_t) \right ) \nabla_\phi V_\phi(s_t) \right].
\end{equation}
To obtain $\widehat{V}(s_t)$, we typically use a $K$-step bootstrap version $\widehat{V}(s_t) = r_t + \gamma r_{t+1} + \gamma^2 r_{t+2} +\ldots + \gamma^{K-1}r_{t+K-1} + \gamma^K V_\phi(s_{t+K})$. Note that although the expression for  $\widehat{V}(s_t)$  looks the same as that for $\widehat{Q}(s_t,a_t)$ in \cref{eq:hatQK}, the conditions of sampling are different: $\widehat{V}(s_t)$ is used for approximating the value of state $s_t$ regardless of action $a_t$ to be taken. To evaluate the gradient $\nabla_\phi {L}(\phi)$, we use a similar approach that we used to estimate the policy gradient in \cref{eq:mc_pgrad_values}; we replace the expectation with a Monte Carlo estimate obtained from one or multiple trajectories.

\begin{figure}[!t]
\centering
\includegraphics[scale=1]{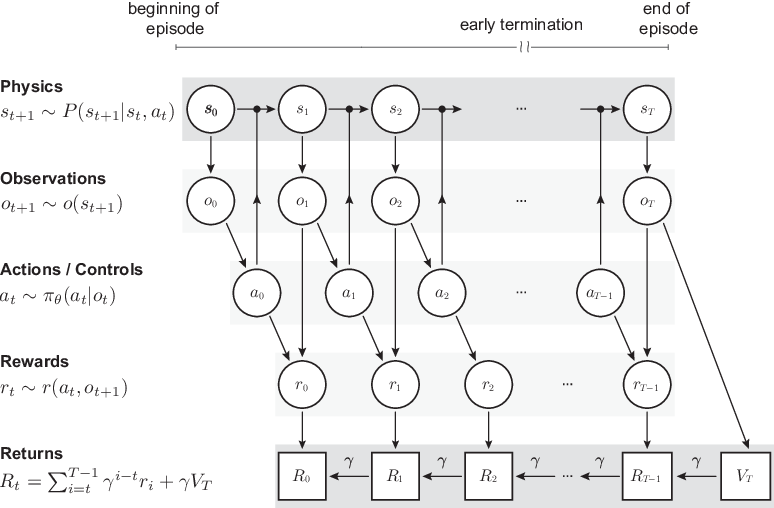}
\caption[Data generation
during a typical model-free RL training episode.]{\textbf{ 
Data generation during a typical model-free RL training episode.} Physics determines how the true state $s_t$ of the body and the world evolves, and observations $o_t$ are the agent-observable elements of the physics. Actions $a_t$ are determined by the agent's policy $\pi_\theta$. The policy could retain a memory of previous observations and actions not shown in this figure. The sequence of immediate rewards $r_t$, while assigned based on the resulting physical state in our model, can in general depend on the previous action taken and the current observations according to the agent's interpretation. Estimated infinite-horizon returns $\widehat R_t$ are then computed from the discounted future rewards plus the agent's expected return estimated by the value function $V_T = V_\phi(o_T)$ after the available storage of rewards is exhausted. %
The objective of the learning agent is to maximize the expected value of return over all possible trajectories.}
\label{fig:flow}
\end{figure}

The choice of the $Q$-function approximator is not unique: the $K$-step truncated approximator in~\cref{eq:qvalKstep} and the $K$-step bootstrap approximator in~\cref{eq:hatQK} are two examples. Alternatively, one could choose to represent the $Q$-function by a parametric function approximator $Q_\phi(s_t,a_t)$ and train directly from sampled trajectories~\cite{Silver2014,Fujimoto2018}. 
An important consideration in choosing the $Q$-function estimator
is its influence on the bias and variance of the associated gradient estimate in~\eqref{eq:mc_pgrad_values}. 
The variance of~\eqref{eq:mc_pgrad_values} is typically reduced when using the bootstrap estimator in \cref{eq:hatQK} and choosing a smaller $K$ and a larger $\gamma$.
In contrast, using the truncated estimator in~\cref{eq:qvalKstep} and choosing a larger $K$ and smaller $\gamma$ will induce lower bias but higher variance.
The choice of the baseline estimate $\widehat{V}(s_t)$, as long as it does not depend (directly or indirectly) on the action, does not affect the bias of the gradient estimate. 
Besides controlling bias and variance of the gradient estimate, the use of approximations such as in \cref{eq:qvalKstep} and \cref{eq:hatQK} relieves us from the impractical requirement of having to collect very long trajectories to approximate the gradient estimate. Instead we can use short trajectory segments or even individual transitions to compute a gradient estimate. When $K=1$, the update rule obtained from the one-step truncated estimator is equivalent to the well-known temporal difference (TD) learning method \cite{Sutton1988, Sutton2018}.  %

In partially-observed environments, instead of knowledge of the full state $s_t$, the agent has access only to partial observations $o_t$.  We thus have to replace $s_t$ with $o_t$ as the input to the $V$ and $Q$ functions. Doing this violates the mathematical rigor that justifies the replacement of the expected return $R_t$ in \cref{eq:pgrad} by the advantage $\mathcal{A}_t = Q(o_t,a_t)-V(o_t)$ in~\cref{eq:pgrad_values}, and may lead to situations where two states with different expected returns have the same advantage. Nonetheless, with proper choice of observations, a direct implementation of actor-critic methods to POMDP settings has proved effective for solving a variety of problems~\cite{Verma2018,Novati2018b,Buzzicotti2021}.

\section{Training the RL agent}
\label{sec:rl_train}
RL training is often structured into \textit{episodes}, which correspond to short bouts of interaction with the environment. In each episode, we specify an initial configuration for the body and initial state for the environment, perform a policy \textit{rollout}, and terminate the episode under specific conditions that indicate success, failure, or time limit. After the termination, we reset the body and the environment to start the next episode. 
The training process within an episode is summarized in~\cref{fig:flow}. 

At the beginning of a new episode, the environment state is initialized to $s_0$, and a set of initial observations $o_0$ is provided to the agent to produce the initial actions $a_0$. As time evolves, a new set of states $s_{t+1}$ is generated according to the transition rule $P(s_{t+1}|s_t,a_t)$, and new observable features $o_{t+1}$ of the environment are generated accordingly. The agent then uses its control policy $\pi_\theta(a_t|o_t)$ to decide what new action $a_{t+1}$ to take based on the new observations $o_{t+1}$, possibly in conjunction with its own internal memory. The reward $r_t$ is assigned based on the currently available variables, including state, observation, and actions, and directly provided to the agent; endogenously-generated rewards are computed based on the agent's observations and potentially memory. The collected rewards $r_t$ are then used to calculate a return ${R}_t$ from the policy rollout at each time step in the training episode. 
The expected return obtained from running a number of training episodes defines an objective function, as formalized in \cref{sec:rl_math} and \cref{sec:Training:AC}, and is used to update the parameters $\theta$ of its control policy $\pi_\theta(a_t|o_t)$ (\cref{fig:intuition}). 

\subsection{Agent Parameterization}
\label{sec:parameterization}
To represent the control policy (actor) $\pi_\theta$ and value function (critic) $V_\phi$,  neural networks are the preferred choice of function approximators.
Parameterizing the action distribution of a policy using neural networks is a decades-old idea \cite{Williams1992, Miller1995, Lewis1996}, and is closely related to the reparameterization trick more recently re-popularized in the context of variational autoencoders \cite{Kingma2013, Rezende2014}.  
Parameterizing probability densities using neural networks allows for gradient-based optimization of probability distributions; see \cite{Schulman2015a} for a unified interpretation of computing gradients along stochastic computation graphs. 
In the simplest case, the policy and the value function would be parameterized by separate neural networks, but architectures allowing for sharing intermediate representations are relatively common \cite{Mnih2016}, where the parameters $\theta$ and $\phi$ are often lumped into a single notation $\theta$ for both policy and value function. 

A stochastic policy is represented by a neural network whose outputs can be interpreted as the parameters of the probability distribution from which the actions are sampled. For a control problem with continuous actions, the policy is often chosen to follow a Gaussian distribution, where the neural network determines the mean and variance of the Gaussian distribution; namely,~a policy $\pi_\theta(a_t|s_t)$ would be implemented as $\mathcal{N}(a_t|\mu_\theta(s_t), \sigma_\theta(s_t))$, where the mean $\mu_\theta(s_t)$ and variance $\sigma_\theta(s_t)$ are given as outputs of the neural network. It is then possible to take gradients with respect to policy by backpropagating through the layers of the neural network that parameterizes $\pi_\theta(a_t|s_t)$. Policies with different probability distributions, such as those in the exponential family, can be similarly represented. If a policy is required to provide both discrete and continuous actions, the output of the neural network representing this policy would partly correspond to parameters of binomial or multinomial distributions and partly to continuous distributions.

\subsection{Agent Updates}
\label{sec:agent:updates}
We discuss how to train and update the agent
using the popular TRPO \cite{Schulman2015b} and PPO \cite{Schulman2017b,Heess2017} algorithms as examples.
Although the use of good critic estimators helps to modulate the variance of the policy, gradient estimates computed according to \cref{eq:mc_pgrad_values} could still suffer from high variance due to the noise inherent to individual trajectories. 
We can mitigate this problem by averaging over more rollouts, i.e.,~larger $M$ in \cref{eq:mc_pgrad_values}.
However, literal adherence to \cref{eq:pgrad} and \cref{eq:mc_pgrad_values} means we can only perform a single update of the policy after collecting $M$ rollouts, because we require trajectories sampled from the \emph{current} policy to approximate the expectations with respect to $\pi_\theta$. In practice, this leads to data-inefficient algorithms. A number of techniques have been developed to address this problem. These are broadly referred to as \emph{off-policy} learning techniques.
\emph{Off-policy}, in contrast to \emph{on-policy}, refers to the use of trajectory data that was produced by a policy different from the one that is currently being updated. 
These techniques typically either rely on value functions, or use importance sampling~\cite[Chapter~9]{Owen2013,Tokdar2010} to correct for the mismatch in the sampling distribution, or both.

Here, we briefly summarize the approach taken in \cite{Schulman2015b,Schulman2017b} as an example of a technique that allows limited data reuse. To understand the algorithms in~\cite{Schulman2015b,Schulman2017b}, it is helpful to note that we can obtain the gradient estimates in \cref{eq:mc_pgrad_values} by differentiating a ``surrogate'' objective $\mathcal{J}^{\mathrm{surr}} = \mathbb{E}_{\pi_\theta} [ \widehat{\mathcal{A}}(s_t,a_t)]$ along a trajectory, by choosing an estimate $\widehat{\mathcal{A}}(s_t,a_t)$ of ${\mathcal{A}}(s_t,a_t) = Q(s_t,a_t) - V(s_t)$ but ignoring its dependence on $\pi$ when computing the gradient $\nonumber\nabla_\theta \mathcal{J}^{\mathrm{surr}}(\theta)$. Namely, using the estimate in \cref{eq:hatQK}, we get
\begin{equation}
    \widehat{\mathcal{A}}(s_t,a_t) %
    = r_t + \gamma r_{t+1} + \gamma^2 r_{t+2} + \dots + \gamma^{K} V_\phi(s_{t+K}) - V_\phi(s_{t}).
\end{equation}
Realizing that the surrugate objective can be written as
\begin{align}
    \mathcal{J}(\theta) =& \mathbb{E}_{\pi_\theta} \left [ \widehat{\mathcal{A}}(s_t,a_t) \right ] = \mathbb{E}_{\pi_{\theta_\mathrm{old}}} \left [ \frac{\pi_\theta(a_t|s_t)}{\pi_{\theta_\mathrm{old}}(a_t|s_t)} \widehat{\mathcal{A}}(s_t,a_t) \right ],
    \label{eq:J_imp_sampled}
\end{align}
we can approximate the Monte-Carlo estimate of the gradient in \cref{eq:mc_pgrad_values} using
\begin{equation}
\label{eq:mc_grad_is}
\nabla_\theta \mathcal{J}(\theta)
    \approx \frac{1}{M} \sum_{j=1}^M
    \frac{\pi_\theta(a_t^j|s_t^j)}{{\pi_{\theta_\mathrm{old}}}(a_t^j|s_t)} \widehat{\mathcal{A}}(s^j_t,a^j_t) \nabla_\theta \log \pi_\theta(a_t^j|s_t^j),
\end{equation}
where $\theta_\mathrm{old}$ are the parameters \emph{before} the update. This correction constitutes only a partial solution since $\widehat{\mathcal{A}}$ also depends on $\pi$; see, \textit{e.g.}, \cite{Espeholt2018} for an algorithm that also corrects for changes in $\widehat{\mathcal{A}}$ by employing importance sampling more extensively. For this reason, and more generally because gradient estimates obtained according to \cref{eq:mc_pgrad} and \cref{eq:mc_grad_is} can lead to uncontrolled, large changes even when averaged over many trajectories, it is often desirable to \emph{limit the change in the policy} that we allow from one iteration to the next. This intuition can be implemented in different ways. The algorithms described in \cite{Schulman2015b, Schulman2017b} use the \emph{Kullback-Leibler divergence} \cite{Kullback1997} to measure the difference $\mathrm{KL}[\pi_{\theta_\mathrm{old}} || \pi_\theta]$ between $\pi_\theta$ and $\pi_{\theta_\mathrm{old}}$ and optimize \cref{eq:mc_grad_is} subject to the constraint that $\mathrm{KL}[\pi_{\theta_\mathrm{old}} || \pi_\theta] < \delta$. This strategy is referred to as a \emph{trust-region} method and can greatly improve the stability of policy optimization algorithms; see \textit{e.g.}\ \cite{Peters2010, Abdolmaleki2018, Heess2015b} for further examples.

Putting together the ideas in~\cref{sec:rl_math}--\cref{sec:rl_train}, we arrive at an algorithm of model-free reinforcement learning in an actor-critic style as illustrated in \cref{alg:mfrlac}; See \cref{app:ppo} for a more detailed description of PPO algorithm.

Lastly, exploration can significantly affect the learning process~\cite{Thrun1992}.
One practical way of achieving exploration beyond what the stochastic policy offers is to inject noise into the policy \cite{Silver2014, Lillicrap2015, Mnih2015}. That is, we add an entropy bonus term in the objective function to guarantee sufficient exploration. This augmented objective function ensures that the agent succeeds at the given task while acting as randomly as possible. This exploration method was originally proposed by \cite{Williams1991} and is widely used in different learning algorithms~\cite{Mnih2016, Schulman2017b, Haarnoja2018}; see \cref{app:ppo} for the usage of this method in PPO.

\begin{algorithm}[t]
\caption{Model-Free Reinforcement Learning with Actor-Critic Methods}
\label{alg:mfrlac}
\begin{algorithmic}[1]
\For{time step $t = 0,1,...$ }

\If{$t=0$ \textbf{or} previous time step is end of an episode} 
 \State reset environment state $s_t\sim P(s_0)$ and return an initial observation $o_t\sim o(s_{t})$
\EndIf{}
\State sample action from policy using current observation: $a_t \sim \pi_\theta(a_t|o_t)$
 \State step forward environment states using sampled action: $s_{t+1}\sim P(s_{t+1}|s_t,a_t)$
 \State obtain new observations $o_{t+1}\sim o(s_{t+1})$ and evaluate the reward function $r_t \sim r(a_t,o_{t+1})$
 \If{stored trajectories are enough for an update}
 \For{update epoch number $k = 0,1, \hdots K$}
 \State estimate expected return at each time step of stored trajectories $\widehat{Q}_t $
 \State evaluate value function at each time step $V_t = V_\phi(s_t)$
 \State compute the advantage at each time step  $\widehat{\mathcal{A}}_t = \widehat{Q}_t - V_t $
 \State compute the loss functions \cref{eq:bellman_loss} and  \cref{eq:J_imp_sampled} using advantage and value estimates
 \State update parameters of policy and value function to minimize each loss accordingly
 \EndFor
 
 \EndIf
 
\EndFor

\end{algorithmic}
\end{algorithm}

\section{Modeler Choices}
\label{sec:implement}

In setting up an embodied learning problem, we have to properly select and model the observations, actions, and rewards underlying a desired behavior without sacrificing the physical or biological meaning of the learning problem.
Here, we give an overview of best practices in designing the RL problem following the order of each item's appearance during a rollout as shown in both \cref{fig:flow} and \cref{alg:mfrlac}. 

\subsection{Episode Initialization} 
To ensure that data collected during every episode productively contributes to learning the task, one must appropriately select the initialization distribution $P(s_0)$.
During the early stages of training, most of the agent's experience comes from the portion of state space near these initial states, as the agent has not yet learned to produce effective behavior for the domain. Therefore, the environment modeler must should tailor the initialization distribution to encourage the agent encounter states that are likely to be relevant for successful task execution.
In other words, the agent needs to be exposed to a distribution of initial states that enables progressive learning by curriculum, wherein there are states from which learning will be possible for initial policies and immature policies. This initial distribution may even be dynamically adjusted during learning, so that the task difficulty gradually increases as the policy gets improved.

\subsection{Agent-centric Observations}
Observations are functions of the states of the system, including both the environment and the body. Different observations can result in learning problems of dramatically varying difficulty and different types of control policy. In designing the observation function, we must be mindful of a few key considerations.

\paragraph{Degree of partial observability} In certain situations, such as for an agent navigating a complex maze, one might consider that the agent would benefit from a complete map of the environment (\textit{i.e.}, a whole grid world), allowing us to perform classical planning\cite{Lavalle2006}. On the opposite extreme, one can try to train a blind agent without any observations, in which case the agent would have to rely entirely on memorization and path integration.  In between these extremes, we can consider training agents that get partial information; with partial information, navigating a maze goes from something requiring no memory to a task that requires keeping track of routes and localization, and integrating this with sensory information \cite{Mirowski2016, Wayne2018}.  While full-state observations may seem intuitively advantageous, complete information can become overwhelming as the state dimensionality increases. Processing enormous streams of information can be computationally costly, and high dimensional data may hide the important features among many task-irrelevant dimensions.
Indeed, in many biological settings, organisms learn or evolve to operate with just the sufficient amount of information to perform essential tasks well enough \cite{Fonio2016, Heyman2017, Feinerman2018}, so a data-efficient RL agent should strive for the same.

\paragraph{Observation invariance}  Certain behavior should be invariant to some irrelevant observations. For example, the ability to control a body to stand and walk should be independent of the body's absolute position in the world.  A locomotion controller should primarily depend upon proprioception of the relevant body parts and could be designed in such a way as to accept instructions that are separately informed by richer sensory information \cite{Merel2018, Merel2019b, Merel2019a}. This design ensures that essential actions are robust to changes in external conditions.
Recent work has explicitly investigated how egocentric visual observations facilitate generalization through invariance \cite{Hill2019}.

\paragraph{Biological plausibility} To model an embodied agent in an ecologically plausible way, it's important that its perceptions and observations are always egocentric to its body \cite{Gibson2014}. 
Egocentric observations naturally lead to actions invariant under spatial transformations about the body, and can simplify the computational demand to encode complex observables of an environment. For example, egocentric vision makes it easier to maintain consistent dimentionality in observations while describing scenes with a variable number of objects, each with other varying properties~\cite{Merel2019c}.
As an aside, if the training process is not meant to exactly parallel ecologically-valid learning, one can choose to supply the value function with additional observations that are not necessarily plausible for a realistic agent to obtain, while keeping inputs to the policy ecologically sound. This method ensures that, at test time, the policy is biologically meaningful, even if it was trained using privileged information to speed up and stabilize training.

\paragraph{Simplicity for function approximation} Since function approximators such as neural networks would be used to solve the learning problem, the choice of observation representation will have impact on the difficulty of learning.
For example, if we want to learn a control policy based on input of a robot joint angle $\beta$ and the ideal controller is a linear function of $[\sin(\beta), \cos(\beta)]$, using both $\sin(\beta)$ and $\cos(\beta)$ as the observations will make learning process much simpler than using the value of $\beta$.
While representation learning from raw inputs is a strength of deep learning, the rates of learning, as well as the space of learnable tasks with a limited-capacity neural network, may depend on the choice of input observations. %

\paragraph{Goal-instructed behaviors} In a multi-task scenario, additional observations that serve as instructions are routinely employed. For example, in navigating to a target sampled from a distribution of locations, the coordinates or an image of the target may be provided as an input \cite{Tassa2014, Zhu2017}. Similarly, if we train an agent to perform multiple distinct tasks, a contextual signal may be provided to indicate which task to complete. Generally, we can either have a higher-level controller provide this information or manually instruct the policy. Arguably more ecological is an agent that solves multiple tasks without explicit instructions to specify the task, but infers the wanted task from the visual observations (\textit{e.g.} \cite{Merel2020}), possibly in conjunction with an assessment of its internal states.

\subsection{The Action Space}
The next modeling choice is to decide how the action produced by the agent controls the physical environment. For example, when controlling a single motor, the agent may produce actions within a convenient range, and this action is interpreted as a control signal which maps to the available range of motor activity. In practice, it may make a problem easier if all actions are normalized to be within a reasonable range relative to each other. Furthermore, different protocols exist for the control of an actuator, \textit{e.g.}, a joint can be driven either by position-control, torque-control, or even muscle-like tensile elements \cite{Millard2013, Angles2019, Marjaninejad2019, Lee2019}. That is, for a motor to control a joint with a limited range of motion, the control signal could specify the target position of the joint, or the torque to apply in either direction. In conventional robotics, torque-based control is often perceived as more fundamental, though in a learning context one may care more about which actuation scheme enables faster and more robust learning instead of the detailed realizability as low-level controllers \cite{Peng2017}.
In other situations where a known low-level controller is preferred, one could also simplify learning by asking the control policy to output discrete actions such as push or pull, instead of the amount of force to be applied to an extensile element.

\subsection{The Reward Function}
The reward, whether designed by the engineer or scientist or innately specified through evolution in animals, induces the emergence of the rewarded behavior.  
We often need to engineer the reward function to induce the agent to learn a desired behavior. A realistic workflow is that we first propose a naive reward function with a particular goal, obtain a policy that can achieve high rewards, and then reverse-engineer the reward function iteratively until the policy produces the desired behavior we have in mind. Such an iterative procedure is sometimes necessary because the designer of the reward, based on their strong intuitions with real-world physics, can easily neglect components of the reward or overlook the possibility of unintended behaviors that can obtain decent rewards. 

Rewards can be dense or sparse, both with respect to time and space. 
Sparse rewards are provided only when a goal is achieved, remaining zero otherwise. In contrast, a dense reward means some nonzero value being provided at each time step. Sparse rewards are easy to program but harder to learn since the agent gets no signal other than the time step when it reaches the goal. 
In such settings, exploration of the environment is critical. Learning will stagnate if the distribution of states explored by the agent does not overlap with the rewarding ones with a high enough probability. Thus for a task, such as moving to a goal, we may define a dense ``shaping'' reward to provide a broader learning signal. A convenient way to generate normalized shaping rewards is to specify a metric between the goal and an arbitrary state, then set the reward function to be the exponential of the negative distance between the goal and the agent's current state. For example, when trying to navigate towards a reference pose $\mathbf{x}_{\rm ref}$, a dense reward could be defined as $\exp[-\gamma\|\mathbf{x}_{\rm ref} - \mathbf{x}_{\rm current}\|^2]$ where $\gamma$ controls the steepness with which the reward falls off when the agent's current position deviates from the reference. In contrast, the sparse version of this reward would be to provide $1$ if and only if the $\|\mathbf{x}_{\rm ref} - \mathbf{x}_{\rm current}\|^2 < \epsilon$ for some small $\epsilon$ and $0$ otherwise.

\subsection{Termination Criteria} 
An episode is commonly terminated after a predetermined duration or upon the success or failure of certain conditions. 
This aspect of the training can be critical for success in specific problems as well-designed episode resets can be quite beneficial, much like when parents pick their children up as they fall, so the children spend more time walking instead of crying on the floor. However, careful attention is needed in setting termination conditions to prevent undesirable edge cases, such as an agent lingering just outside a termination-triggering area and collecting low rewards, or an agent committing suicide after finding an easy way to trigger termination to avoid temporary negative rewards.
Moreover, it is usually essential to distinguish different types of terminations when calculating returns. We should use value function bootstrapping in regular terminations (\textit{e.g.}, an episode reaching its time limit) and successful terminations (\textit{e.g.}, the agent reaching the goal) and use truncation in cases of failure terminations to further disincentivize such terminating states.

\subsection{Decision Time Step} 
Another concern for setting up a training episode is choosing the correct step size for both the physical simulation and the decision process.
While many physical phenomena can be conceived as evolving continuously over time, they are often simulated on a computer at some spatial and temporal discretization or sampled from high-speed photography at a given framerate. %
We thus need to consider the appropriate relation between the physics step size and our decision step size. Typically, the temporal resolution of the agent control should be coarser than that of the physics time step. The decision timestep should reflect the reaction time of an animal's nervous system or a robot's information transmission system. Alternatively, a delay from the observation can be introduced to the action to mimic the reaction time. The learned policies may be unrealistic if such measures are not implemented carefully. For example, if an animal is modeled with too short of a reaction time, it might learn to dodge predators in close quarters instead of trying to interpret subtle long-range sensory cues to avoid all encounters. Although one strategy is not necessarily superior to the other, a well-designed learning setup should only allow such ambiguities if explicitly desired.

\section{Conclusions}
Recent advances in simulations, robotics, and machine learning have made it possible to explore the sensing and control laws that guide the behavior of complex bodies interacting within rich environments~\cite{Bruce2024}. Indeed, the integration of learning algorithms with simulations or robotic experiments offers unprecedented opportunities for advancing our understanding of physical intelligence~\cite{Sitti2021} and animal behavior~\cite{Verma2018,Hannigan2019,Gunnarson2021, 
Singh2023}, and for developing autonomous robots that learn from physical interactions with the environment much like their biological counterparts~\cite{Choi2023, Li2023, Liang2024}.

This work presented a concise exposition of the mathematical and algorithmic aspects of reinforcement learning methods as a tool for investigating animal and robot behavior. We explained the subtlety in each component of embodied RL and the critical design considerations to formulate such a problem. With this understanding, the reader is now ready to explore the plethora of advanced RL algorithms, including those that consider other types of function approximators, such as recurrent neural networks~\cite{Ni2021,Kapturowski2018,Song2018} and transformers~\cite{Radosavovic2024,Chen2021,Yang2021}, and incorporate large language models~\cite{Pternea2024,Xie2023,Katara2023,Ouyang2024}. Importantly, this exposition serves as the foundation to pursue innovative applications in robotics and at the interface of biology, physics, and computational neuroscience.

\section*{Acknowledgments}
Kanso would like to acknowledge support 
from the Office of Naval Research (ONR) Grants N00014-22-1-2655, N00014-19-1-2035, N00014-17-1-2062, and N00014-14-1-0421, and the National Science Foundation (NSF) Grants RAISE IOS-2034043 and CBET-2100209.
The authors are thankful for useful discussions with Dr. Yuval Tassa.

\bigskip

\bibliographystyle{vancouver}
\bibliography{references}

\appendix

\section{Proof for the derivation of objective function}
\label{sec:appendix:obj}
We show the proof of 
\begin{equation}
\label{eq:pgrad_zero}
\mathbb{E}_{\pi_\theta}\left[\sum_{t=0}^{\infty}\sum_{t'=0}^{t-1} \gamma^{t'}r_{t'}\nabla_\theta \log\pi_\theta(a_t|o_t) \right] = 0
\end{equation} 
to arrive at \cref{eq:pgrad}.
Consider a single component in the summation
\begin{equation}
\label{eq:appendix:prove_exp}
    \mathbb{E}_{\pi_\theta}\left[\gamma^{t'}r_{t'}\nabla_\theta \log\pi_\theta(a_t|s_t) \right].
\end{equation}
This expectation is taken with respect to all trajectories that pass states $s_t,s_{t'}$ and actions $a_t,a_{t'}$ at time $t,t'$, respectively, and follows the policy. We can rewrite the expression \cref{eq:appendix:prove_exp} as
\begin{equation}
\label{eq:appendix:prove_prob}
    \int_{s_t,a_t,s_{t'},a_{t'}} p_\theta(s_t,a_t,s_{t'},a_{t'})\gamma^{t'}r_{t'}\nabla_\theta \log\pi_\theta(a_t|s_t),
\end{equation}
where $p_\theta(s_t,a_t,s_{t'},a_{t'})$ is the probability density function (PDF) of trajectories that include $s_t,a_t,s_{t'},a_{t'}$ and follow the policy $\pi_\theta$. Then using the chain rule of probability, we can rewrite the PDF $p_\theta(s_t,a_t,s_{t'},a_{t'})$ as $p_\theta(s_t,s_{t'},a_{t'})p_\theta(a_t|s_t,s_{t'},a_{t'})$, and \cref{eq:appendix:prove_prob} becomes
\begin{equation}
\label{eq:appendix:prove_cond}
    \int_{s_t,s_{t'},a_{t'}} \left[p_\theta(s_t,s_{t'},a_{t'}) \gamma^{t'}r_{t'}\int_{a_t}p_\theta(a_t|s_t,s_{t'},a_{t'})\nabla_\theta \log\pi_\theta(a_t|s_t)\right].
\end{equation}
Now if $t'<t$, meaning $(s_{t'}, a_{t'})$ happens before $s_t$, then $a_t$ is independent of $s_{t'}$ and $a_{t'}$, so we have $p_\theta(a_t|s_t,s_{t'},a_{t'}) = \pi_\theta(a_t|s_t)$. The inner integral in \cref{eq:appendix:prove_cond}
\begin{align}
\label{eq:appendix:pg_result}
    &\int_{a_t}p_\theta(a_t|s_t,s_{t'},a_{t'})\nabla_\theta \log\pi_\theta(a_t|s_t)= \\
    \nonumber& \hspace{1.5in}\int_{a_t}\pi_\theta(a_t|s_t)\nabla_\theta \log\pi_\theta(a_t|s_t)= \nabla_\theta\int_{a_t}\pi_\theta(a_t|s_t)
    =0.
\end{align}
Here we use the ``likelihood ratio trick'' introduced in \cref{sec:rl_math} in the second equality. Therefore, we can conclude that \cref{eq:appendix:prove_exp} is always zero for $t'<t$, and \cref{eq:pgrad_zero} has been proved. Note that in the other case where $t'>t$, the probability density of $a_t$ is impacted on the future state $s_{t'}$ in the conditional PDF $p_\theta(a_t|s_t,s_{t'},a_{t'})$, so we cannot come to the same result as \cref{eq:appendix:pg_result}.

\section{Policy gradient with value functions}
\label{sec:appendix:pg}
We show how the gradient in \cref{eq:pgrad} can be expressed in terms of the advantage $A(s,a) = Q(s,a) - V(s)$ of $\pi$ in~\cref{eq:pgrad_values},
\begin{align*}
    \nabla_\theta \mathcal{J}(\theta)
    &= \mathbb{E}_\pi\left[ \sum_{t=0}^{\infty} \gamma^t R_t \nabla_\theta \log \pi_\theta(a_t|s_t) \right], \\
    & = \mathbb{E}_\pi\left[ \sum_{t=0}^{\infty} \gamma^t \mathbb{E}_\pi \left [ R_t | s_t, a_t \right ] \nabla_\theta \log \pi_\theta(a_t|s_t) \right], \\
    & = \mathbb{E}_\pi\left[ \sum_{t=0}^{\infty} \gamma^t Q(s_t,a_t) \nabla_\theta \log \pi_\theta(a_t|s_t) \right], \\
    & = \mathbb{E}_\pi\left[ \sum_{t=0}^{\infty} \gamma^t \left( Q(s_t,a_t)-V(s_t)\right) \nabla_\theta \log \pi_\theta(a_t|s_t) \right]
    &= \mathbb{E}_\pi\left[ \sum_{t=0}^{\infty} \gamma^t  A(s_t,a_t)\nabla_\theta \log \pi_\theta(a_t|s_t) \right],
\end{align*}
where the first equality follows from the law of iterated expectations, the second uses the definition of $Q$ in \cref{eq:qval}, and the last line makes use of the following argument
\begin{align*}
    \mathbb{E}_\pi\left[ V(s_t) \nabla_\theta \log \pi_\theta(a_t|s_t) \right] &= \int \pi(a_t|s_t) V(s_t) \nabla_\theta \log \pi_\theta(a_t|s_t) \mathrm{d} a_t, \\
    &= V(s_t) \int \pi(a_t|s_t)  \nabla_\theta \log \pi_\theta(a_t|s_t) \mathrm{d} a_t, \\
    &= V(s_t) \int \pi(a_t|s_t) \frac{\nabla_\theta \pi_\theta(a_t|s_t)}{ \pi_\theta(a_t|s_t)} \mathrm{d} a_t , \\
    &= V(s_t) \int \nabla_\theta \pi_\theta(a_t|s_t) \mathrm{d} a_t , \\
    &= V(s_t) \nabla_\theta \underbrace{\int  \pi_\theta(a_t|s_t) \mathrm{d} a_t}_{\mathrm{const.}} = 0.
\end{align*}
In fact, this result suggests that instead of $V$ we can choose \emph{any} function that does not depend on $a_t$ (or the future trajectory) as the baseline. The role of the baseline is to reduce the variance when the gradient is approximated with samples. For this role,   $V$ is not the optimal but an intuitive and convenient choice \cite{Greensmith2004}.

\section{Proximal Policy Optimization (PPO) Algorithms}
\label{app:ppo} We review the Proximal Policy Optimization (PPO) method \cite{Schulman2017b}. Specifically, we focus on one form of this method known as (soft) clipped PPO.
This algorithm ensures fast learning and robust performance by limiting the amount of change for the policy between successive updates.

The algorithm uses a surrogate objective (see section \cref{sec:agent:updates}) that gets clipped when the policy deviates too much from the reference policy from the previous cycle:
First, a copy of the policy $\pi_{\theta}$, denoted by $\pi_{\theta_{\textrm{old}}}$ is kept unchanged during each agent update cycle. Next, a surrogate objective function is defined as
\begin{equation}
    \mathcal{J}_{\text{clip}}(\theta,\phi)  = \mathbb{E}_{t{\textrm{ in a cycle}}}\left[\min\left(A_t \cdot\dfrac{\pi_\theta(a_t|o_t)}{\pi_{\theta_{\textrm{old}}}(a_t|o_t)}, A_t \cdot \operatorname{clip}
    \left(\dfrac{\pi_\theta(a_t|o_t)}{\pi_{\theta_{\textrm{old}}}(a_t|o_t)}, \epsilon \right)\right)\right].
\end{equation}
where the clip function cuts off the likelihood ratio between the updated and old policy to remain inside the interval $[1-\epsilon, 1+\epsilon]$.  Numerical experiments suggest that a value of the hyperparameter $\epsilon=0.2$ gives near optimal performance across different RL tasks~\cite{Schulman2017b}.
Note how this surrogate objective function chooses the minimum of the clipped and unclipped objective functions. The clipped objective keeps the likelihood ration inside  the interval $[1-\epsilon, 1+\epsilon]$. Specifically, when an action has an advantage ($A_t>0$), the likelihood ratio is clipped to be at most $1+\epsilon$. Conversely, when an action has a disadvantage ($A_t<0$), the ratio is cut off below the threshold value $1-\epsilon$.
Namely, the surrogate objective is cleverly designed to choose the lower bound on the clipped objective, hence limiting change in the policy only when it improves the overall objective.

Effectively, this prevents an update step from increasing the likelihood of performing an advantageous action too much and vice versa. This encourages exploration relative to maladaptive over-exploitation of the environment and thus makes the algorithm more robust. Even though this is not conventionally done, in practice the estimate of the return can also incorporate importance weights based on the ratio between old and update policy \cite{Munos2016, Espeholt2018} as introduced in \cref{sec:agent:updates}.

In addition to the actor critic objectives, it is also common to introduce an entropy bonus for the policy distribution \cite{Williams1992, Mnih2016}. This entropy term encourages the optimization method to find a policy that can succeed at the given task while acting as randomly as possible. This is related to the principle of maximum entropy, which states that given prior data, the probability distribution that best represents the current state of knowledge is the one with largest entropy \cite{Schulman2017b, Shi2019, Ahmed2018}. So finally our surrogate objective function reads
\begin{equation}
\mathcal{J}_{\text{sac}}(\theta,\phi) = \mathbb{E}_{t{\textrm{ in a cycle}}}\left[\mathcal{J}_{\text{clip}}(\theta) - \frac12 (R_t-V_t)^2 + \alpha H(\pi_\theta)\right],
\label{eq:objective_ppo}
\end{equation}
where $\alpha$ controls the relative weighting between the entropy bonus $H(\pi_\theta)$ and the actor-critic objectives. This relative weighting is usually chosen to be very small ($\alpha = 0.01$ in \cite{Schulman2017b}). Maximization of the objective function in \cref{eq:objective_ppo} gives us the (soft) clipped advantage Proximal Policy Optimization (PPO) method \cite{Schulman2017b}.

Next, we discuss the PPO algorithm in two parts: (i) the environment simulation, and (ii) updating the agent parameters. We highly recommend the readers to go to our Github repository~\url{https://github.com/mjysh/RL3linkFish} for a concise implementation of PPO and compare it with the pseudocode provided below.

\subsection{Environment simulation}
This part of the algorithm simulates the environment using action sequence $a_t$ generated by the agent, and stores the observed rollouts for agent updates at every $N$ time steps according to \cref{alg:sim}. We use $n_o$ and $n_a$ to indicate the dimension of observable states and actions.

\begin{algorithm}[t]
\caption{Environment Simulation}\label{alg:sim}
\begin{algorithmic}[1]

\For{time step $t = 0,1,...$ }

\If{$t=0$ \textbf{or} episode terminates} 
 \State store time step of episode termination,
 \State reset state $s_t\sim P(s_0)$
 \State evaluate observation: $o_t \sim o(s_{t})$
\EndIf{}
\State sample action from policy $a_t \sim \pi_{\theta}(a_t|o_t)$ 
 \State evolve next state according to physics $s_{t+1} \sim P(s_{t+1} |s_t , a_t)$ 
 \State evaluate next observation $o_{t+1} \sim o(s_{t+1})$ and reward $r_t \sim r(a_{t},o_{t+1})$
 \If{$t=0$ \textbf{or} mod$(t,N)\neq0$}
 \State append current action, observation, reward, and probability of sampling the action to assemble vectors $a_{N\times n_a}, o_{N\times n_o}, r_{N \times 1}$, and $\pi_{\theta_{\textrm{old}}}(a|o)_{N\times 1}$
 \Else
 \State update agent networks according to \cref{alg:ppo}
 \EndIf
 
\EndFor

\end{algorithmic}
\end{algorithm}

\begin{algorithm}[t]
\caption{Updating the Agent Networks}\label{alg:ppo}
\begin{algorithmic}[1]
\For{update epoch number $k = 0,1, \hdots K$}
\State compute the truncated return using rewards $r_{N\times 1}$ and assemble into vector $R_{N \times 1}$
\State estimate infinite-horizon return using $R_{N \times 1}$ and $V_T=V_\phi(o_T)$ if bootstrapping is desired (see \cref{eq:hatQK})
\State using $o_{N\times n_o}$ and value function $V_{\phi}$, evaluate expected returns at each time step and store into $V_{N \times 1}$
\State compute the advantage $A = R_{N \times 1} - V_{N \times 1}$ and normalize by its mean and variance if desired
\State evaluate the probability of realizing $a_{N\times n_a}$ based on $o_{N\times n_o}$ for the policy $\pi_\theta$, and store to $\pi_{\theta}(a|o)_{N\times 1}$
\vspace*{-0.75em}
\State compute the action-likelihood ratio: $\varrho_\theta= \dfrac{\pi_{\theta}(a|o)_{N\times 1}}{\pi_{\theta_{\textrm{old}}} (a|o)_{N\times 1}}$
\vspace*{0.25em}
\State compute clipped surrogate loss function: 
$\mathcal{L}_{\text{clip}}(\theta) = \operatorname{mean}\left[\min\left[\varrho_\theta\cdot A, \textrm{clip} (\varrho_\theta, 1-\epsilon,1+\epsilon)\cdot A\right]\right]$
\vspace*{0.25em}
\State compute the value-function loss: $ \mathcal{L}_{\textrm{value}}(\phi) = 0.5\cdot\operatorname{mean}\left[(R_{N \times 1} - V_{N \times 1})^2\right]$
\vspace*{0.25em}
\State compute the total loss: $\mathcal{L}(\theta, \phi) = -\mathcal{L}_{\textrm{clip}} (\theta)  + \mathcal{L}_{\textrm{value}} (\phi) - \alpha\cdot \operatorname{entropy}\left[\pi_{\theta}\right]$
\vspace*{0.25em}
\State update parameters $(\theta, \phi)$ to minimize the total loss using a gradient based optimizer (\textit{e.g.}, Adam)
\EndFor
\end{algorithmic}
\end{algorithm}

\subsection{Updating the agent networks}
Every $N$ timesteps, where $N$ is a prescribed number, the parameters of the actor-critic networks within the RL agent are updated for $K$ epochs. The value $K$ is typically chosen to be between $O(1)$ to $O(10^2)$. For simplicity, we assume our continuous action variables follows a multivariate normally-distributed policy $\pi_{\theta}$ with mean value represented by a neural network parameterized by $\theta$ and constant diagonal covariance matrices. To adapt our implementation to a discrete control problem, the policy can be represented by a neural network with a soft-max output layer that gives out the categorical probability of each discrete action directly. The value function $V_{\phi}(o_t)$ is also represented by a neural network with parameters $\phi$.
Then, using the collected trajectories during the last $N$ time steps, the parameters $\theta,\phi$ are updated according a total loss function $\mathcal{L}(\theta, \phi)=-\mathcal{J}_{\text{sac}}(\theta,\phi)$. Details of the computation sequence is included in \cref{alg:ppo}.

There are a number of hyper-parameters involved in this algorithm, the values of which we chose according to the suggestions in \cite{Schulman2017b}; specifically we set $\epsilon = 0.2$, $\alpha =0.01$. However, the optimal values of these hyper-parameters can be specific to the problem one is attempting to solve and we suggest the readers to explore different values for best performance. 

Note that there is another variant of the PPO algorithm, not discussed here, known as PPO-Penalty. This algorithm uses a KL divergence penalty term in the objective function, instead of the clipped surrogate objective, as the soft constraint on how far the updated policy can diverge from the old one. For more details about this variation of the algorithm and its performance we refer the readers to \cite{Schulman2017b, Heess2017}.

Finally, we refer the reader to \href{https://spinningup.openai.com/en/latest/algorithms/ppo.html}{OpenAI's documentation} of the PPO algorithm (and other policy gradient methods) and their \href{https://github.com/openai/baselines/tree/master/baselines/ppo2}{baseline implementations} for a thorough explanation of the details. 

We also refer the reader to several open-source libraries that serve as accessible frameworks with associated suites of tasks that provide rigorous performance benchmarks for RL agents, \textit{e.g.}, OpenAI gym~\cite{Brockman2016}, DeepMind control suite \cite{Tassa2018}, SURREAL~\cite{Fan2018}, RLBench~\cite{James2020}, Jumanji~\cite{Bonnet2023}, Habitat~\cite{Puig2023}.

\end{document}